\documentclass[10pt, a4paper]{article}

\usepackage{lrec-coling2024} 

\usepackage{inconsolata}
\usepackage{graphicx}
\usepackage{multirow}
\usepackage{amsmath}
\usepackage{amssymb}
\usepackage{booktabs}

\title{Factorized Learning Assisted with Large Language Model\\ for Gloss-free Sign Language Translation}

\name{
\begin{tabular}{c}
Zhigang Chen$^{1,2}$, Benjia Zhou$^{3}$, Jun Li$^{1,2}$, Jun Wan$^{1,2,3\dagger}$, Zhen Lei$^{1,2,4}$\\
Ning Jiang$^{5}$, Quan Lu$^{5}$, Guoqing Zhao$^{5}$\\ 
\end{tabular}
} 

\address{
$^{1}$MAIS, Institute of Automation, Chinese Academy of Sciences, Beijing, China \\
$^{2}$School of Artificial Intelligence, University of Chinese Academy of Sciences, Beijing, China \\
$^{3}$Macau University of Science and Technology, Macau, China \\
$^{4}$CAIR, HKISI, Chinese Academy of Sciences, Hong Kong, China \\
$^{5}$Mashang Consumer Finance, Chongqing, China \\
\{chenzhigang2021, jun.wan\}@ia.ac.cn
}

\abstract{
Previous Sign Language Translation (SLT) methods achieve superior performance by relying on gloss annotations. However, labeling high-quality glosses is a labor-intensive task, which limits the further development of SLT. Although some approaches work towards gloss-free SLT through jointly training the visual encoder and translation network, these efforts still suffer from poor performance and inefficient use of the powerful Large Language Model (LLM). Most seriously, we find that directly introducing LLM into SLT will lead to insufficient learning of visual representations as LLM dominates the learning curve.
To address these problems, we propose \textbf{F}actorized \textbf{L}earning \textbf{a}ssisted with \textbf{L}arge \textbf{L}anguage \textbf{M}odel (\textbf{FLa-LLM}) for gloss-free SLT.
Concretely, we factorize the training process into two stages. In the visual initialing stage, we employ a lightweight translation model after the visual encoder to pre-train the visual encoder. In the LLM fine-tuning stage, we freeze the acquired knowledge in the visual encoder and integrate it with a pre-trained LLM to inspire the LLM's translation potential. This factorized training strategy proves to be highly effective as evidenced by significant improvements achieved across three SLT datasets which are all conducted under the gloss-free setting.
 \\ \newline  }

\begin{document}

\maketitleabstract

\section{Introduction}
\label{introduction}
\let\thefootnote\relax\footnotetext{
 $^\dagger$Corresponding author.}

Sign language is the primary form of communication for over 70 million deaf people worldwide. It is a visual language consisting of gestures, body movements, and expressions which has a unique linguistic structure. Therefore, it differs greatly from the natural spoken language. The study of sign language processing can bring great convenience between hearing and deaf people.

\begin{figure}[t]
\centering
\includegraphics[width=1.0\linewidth]{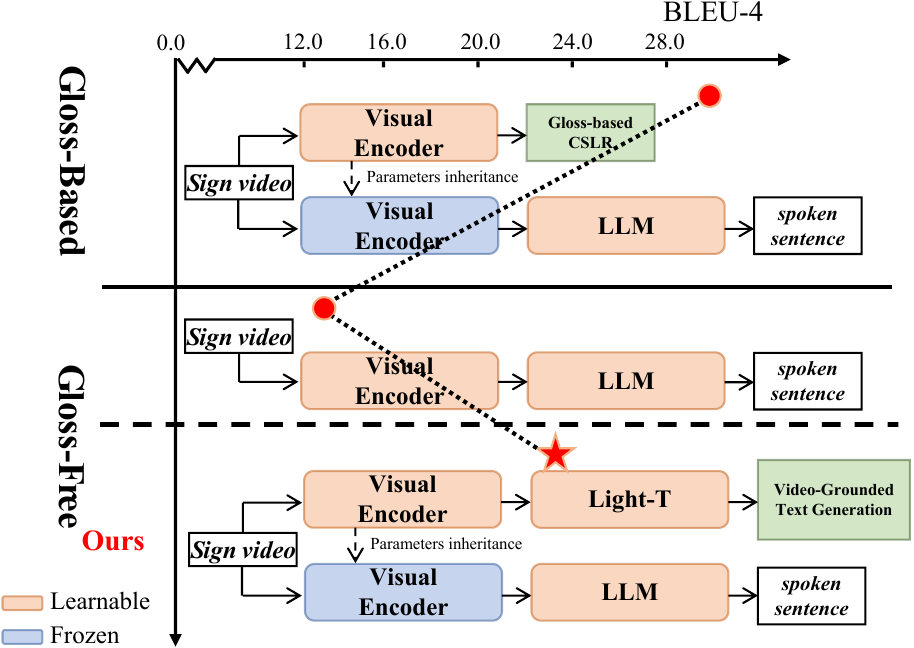}
\caption{Different frameworks and performance of gloss-based and gloss-free methods with LLM. The first row shows the performance of the gloss-based method with LLM~\cite{chen2022simple}. The second and third show the gloss-free method with LLM in our experiments. The BLEU-4 score is gotten on the PHOENIX-2014T test set.}
\label{fig:framwork}
\end{figure}
Unlike Neural Machine Translation~\cite{bengio2000neural}, which focuses on translating between different languages (e.g., English to Chinese), Sign Language Translation (SLT) is a cross-modal task that involves learning visual representations from sign language videos and generating corresponding spoken words. Previous SLT approaches~\cite{camgoz2020sign,zhou2021improving,chen2022simple,chen2022twostream} have relied on gloss sequences to improve performance. Gloss refers to the transcription of signed languages sign-by-sign, where every sign has a unique identifier~\cite{yin2021including}. Gloss sequences are utilized as the supervision for visual representation learning via performing Continuous Sign Language Recognition (CSLR).
Due to the substantial manual labor and specialized linguistic expertise required for gloss annotation, it is challenging to construct large-scale datasets.
Therefore, the existing datasets are relatively small in scale and domain-specific.
Consequently, current gloss-based methods, while achieving good performance on certain specific test sets, suffer from limited generalizability and are unable to benefit from large, high-quality datasets without gloss annotations, such as the newly released large-scale SLT datasets like How2Sign~\cite{duarte2021how2sign} and OpenASL~\cite{shi2022open}.

In summary, the exploration of gloss-free methods is highly necessary, as it can significantly reduce annotation costs and contribute to the development of more reliable and general sign language translation systems.

Recently, there have been attempts~\cite{camgoz2018neural,li2020tspnet,lin2023gloss} to achieve gloss-free sign language translation by jointly training a visual network and translation network in an end-to-end manner. 
To enhance the performance of SLT, the intuitive approach is to employ larger and more powerful pre-trained large language models (LLM) like ~\cite{chen2022simple} (first row in Figure~\ref{fig:framwork}).
However, jointly training the visual encoder and LLM end-to-end without gloss annotations leads to significant performance degradation, as shown in the second row in Figure~\ref{fig:framwork}.

We conjecture it is due to the fact that: 1) the visual encoder was not pre-trained on the sign language dataset leads to poor modeling ability of temporal and spatial information in them. 2) LLM dominated the training process resulting in insufficient learning of visual representations.
We provided a detailed analysis of the specific experiments conducted in Section~\ref{sec:dominance}. This phenomenon has made prior methods use small language models with random initialization resulting in poor performance.
To cope with the above challenges and build a more realistic SLT system, we propose a \textbf{F}actorized \textbf{L}earning \textbf{a}ssisted with \textbf{L}arge \textbf{L}anguage \textbf{M}odel (termed \textbf{FLa-LLM}) for gloss-free SLT. 

Specifically, as illustrated in the third row of Figure \ref{fig:framwork}, we factorize the training process into two distinct stages, visual initialing stage and LLM fine-tuning stage. In the first stage, we introduce a lightweight translation model (Light-T) positioned after the visual encoder to pre-train the visual encoder using a video-grounded text generation task. This strategy can be seen as a soft visual-text alignment method that implicitly supervises the visual encoder with the assistance of a lightweight language translation model, enabling it to acquire language knowledge. While this stage compels the visual encoder to learn semantic visual representations, it may not yield satisfactory translation performance due to the limited strength of the lightweight translation model. To overcome this limitation, we introduce the Large Language Model (LLM) into the second stage namely LLM fine-tuning stage. we incorporate an LLM that has been pre-trained on extensive corpora using an unsupervised approach to enhance the translation performance. The parameters of the pre-trained visual encoder are all frozen to overcome its risk of being biased by the LLM. Finally, we successfully took advantage of LLM in gloss-free SLT and got a BLEU-4 score of 23.09.

In summary, our work makes the following significant contributions:
\begin{itemize}

    \item We analyze the reason why directly training the visual encoder and LLM failed in gloss-free SLT and propose \textbf{FLa-LLM} to overcome this problem. To the best of our knowledge, this is the first successful attempt of LLM on gloss-free SLT.
    \item \textbf {FLa-LLM} method factorizes the training process into two distinct stages namely the visual initialing stage and LLM fine-tuning stage. This division helps mitigate the detrimental effects of the Large Language Model (LLM) on visual representation learning. Moreover, it allows us to leverage the LLM's assistance in SLT at a low cost, improving translation performance.

    \item Our approach greatly boosts the performance of the gloss-free SLT.
    Specifically, we improve the BLEU-4 score by a large margin of 1.65 on PHOENIX14T\cite{camgoz2018neural}, 3.20 on CSL-Daily\cite{zhou2021improving} and 1.63 on How2Sign\cite{duarte2021how2sign} compared with the previous state-of-the-art methods.
\end{itemize}

\section{Related Work}
\noindent \textbf{Gloss-based SLT.} Sign Language Translation (SLT) is proposed by \cite{camgoz2018neural} which intends to translate sign language videos into corresponding spoken sentences. Most SLT methods utilize gloss annotations for pre-training or as assisted supervision which we define as gloss-based SLT. \citet{camgoz2020sign} used the transformer~\cite{vaswani2017attention} architecture and jointly trained Continuous Sign Language Recognition (CSLR) and SLT. \citet{zhou2021improving} utilized gloss as an intermediary and translated spoken sentences back into sign language features to expand the translation corpus. \citet{chen2022simple} transferred a powerful large language model to the sign language domain and improved the performance of SLT significantly. Considering the special structure of sign language, \citet{zhou2021spatial,chen2022twostream} used a multi-cue network for detailed sign language modeling. Gloss-based SLT methods can achieve better performance, but the difficulty of labeling gloss leads to great limitations.

\noindent \textbf{Gloss-free SLT.}
There are a few methods attempting to build a more realistic SLT system with the gloss-free setting. \citet{camgoz2018neural} built an attention-based encoder-decoder framework for SLT. \citet{li2020tspnet} learned hierarchical features of sign language via temporal semantic pyramid. \citet{zhao2021conditional} improved the accuracy and fluency of SLT by conditional sentence generation and cross-modal reranking. \citet{orbay2020neural} utilized adversarial, multi-task, and transfer learning to search for semi-supervised tokenization approaches. \citet{yin2023gloss} analyzed the role of gloss in SLT. Based on it, they proposed gloss attention which enables the model to keep its attention within video segments that have the same semantics locally as the gloss-based model. \citet{lin2023gloss} exploited the shared underlying semantics of signs and the corresponding spoken translation to improve the gloss-free SLT performance. \citet{Zhou_2023_ICCV} integrate contrastive language image pre-training with masked self-supervised learning to create pre-tasks that bridge the semantic gap between visual and textual representations and restore masked sentences. Due to the lack of the assistance of gloss, the above gloss-free methods basically did not use LLM resulting in poor performance of language generation. In addition, the huge computational cost caused by the large number of sign language video frames makes it difficult to perform end-to-end fine-tuning using LLM. 

\section{The Dominance of LLM in SLT}
\label{sec:dominance}
In this section, we analyze the idea presented in Section~\ref{introduction} that LLM dominates the SLT training process when jointly training the visual encoder and LLM end-to-end.

\begin{figure}[t]
\centering
\includegraphics[width=1.0\linewidth]{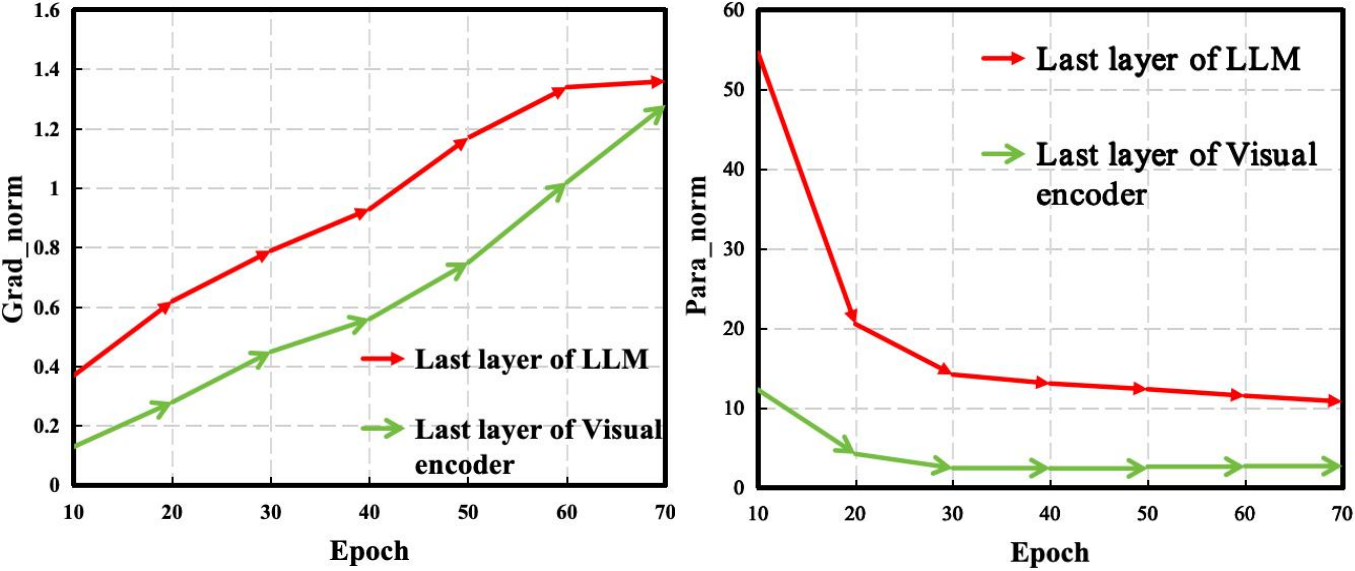}
\caption{The grad norm and parameters norm of the last layer of the visual encoder and the last layer of the LLM when jointly training the visual encoder and LLM end-to-end. }
\label{fig:para norm}
\end{figure}

 The grad norm represents the rate of change of the parameter while the parameter norm represents the magnitude of the change. They can reflect which part of the training process is more active. We chose the last layer of LLM to represent the LLM module and the last layer of the visual encoder to represent the visual encoder. As shown in Figure~\ref{fig:para norm}, we visualized the grad norm and parameters norm of these two layers when jointly training the visual encoder and LLM end-to-end. The grad norm of the last layer of LLM was always greater than the last layer of the visual encoder. At the same time, the parameter norm of the last layer of LLM changed more drastically than the last layer of the visual encoder. It indicates that the main update of the model lies in the LLM module i.e. LLM dominates the SLT training process. It will lead to suppression of visual encoder training and thus failure to learn good visual representations of sign language. The results of Table~\ref{tab:Factorized Learning} experiments similarly prove this statement.

\begin{figure*}[!htp]
  \centering
  \includegraphics[width=1.0
  \linewidth]{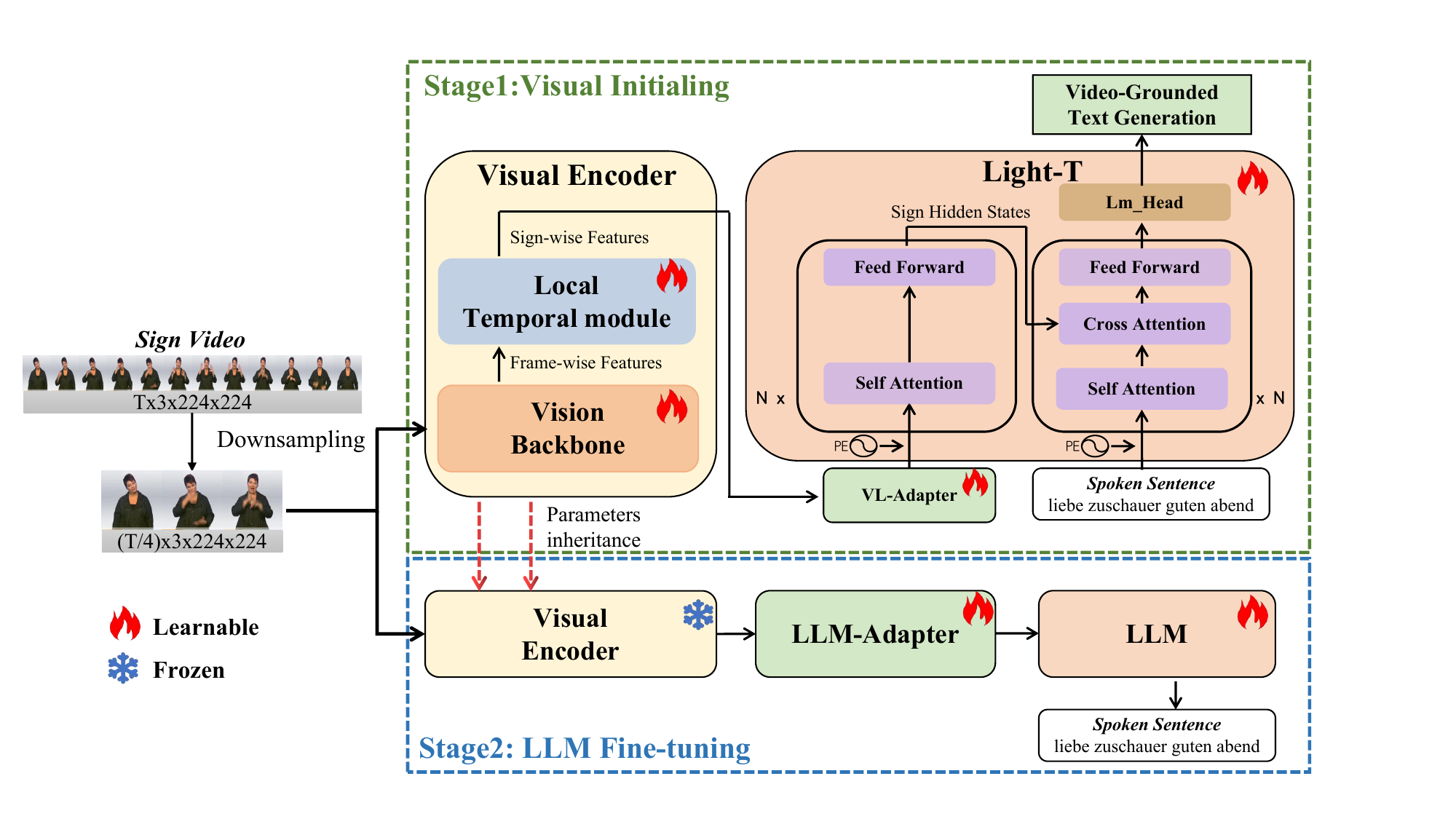}

  \caption{The overall framework of our proposed method. LLM represents the large language model.  }
  \label{fig:Overview}
\end{figure*}
\label{overview}
\section{Method}
\subsection{Overview}
SLT aims to translate a sign video $V=(I_1,I_2,...I_T)$ with $T$ frames into the corresponding spoken sentence $S = (w_1,w_2,...w_U)$ with $U$ words. In this work, we focus on a gloss-free solution that doesn't require gloss annotations. As illustrated in Figure~\ref{fig:Overview}, the training process is factorized into two stages. In the visual initialing stage (Section \ref{visual encoder}), the objective is to facilitate the learning of semantic visual knowledge by the visual encoder from downsampled videos. The visual features are then mapped into a textual embedding space using a Visual-Language Adapter (VL-Adapter). We construct a lightweight translation model (Light-T) to perform video-grounded text generation pre-training. Subsequently, in the LLM fine-tuning stage (Section \ref{LLM}), we retain the pre-trained visual encoder from the visual initialing stage to extract sign-wise features of input videos. Then the features are passed into an LLM-Adapter and LLM to generate corresponding spoken sentences.

\subsection{Visual Initialing}
\label{visual encoder}

\noindent 
Since sign language possesses special spatial properties, the visual initialing of the visual encoder on sign language datasets is necessary. With the gloss-free setting, only the spoken sentences can be used as text supervision.
Therefore, we construct a visual encoder followed by a lightweight translation model (Light-T) to perform visual initialing by a video-grounded text generation task.

\noindent \textbf{Video Downsampling.} The number of frames in a sign language video is greater than the number of words in the corresponding sentence, which has a lot of redundant information.
Therefore, we downsample a $(T \times 3 \times H \times W)$ input sign language video into $(T/4 \times 3 \times H \times W)$ to reduce computational cost without performance degradation.

\noindent \textbf{Visual Encoder.} The visual encoder consists of a vision backbone and a local temporal module. ResNet18~\cite{he2016deep} is chosen as our vision backbone. The downsampled sign video is fed into the visual backbone frame by frame to get frame-wise features. A complete sign language token is often expressed by several continuous frames. Therefore, we designed a temporal module to capture the local timing information within the sign language video. The temporal module consists of one temporal convolutional layer, one batch normalization layer, and one Relu layer. The frame-wise features are passed through the temporal module to get sign-wise features $F = (f_1,f_2,...f_N)$ where $N = T/4 $ with a size of $(T/4 \times C)$. The above operation can be formulated as:
\begin{equation}
    f_{1:N} = \mbox{VisualEncoder}(I_{1:T}). 
\end{equation}

\noindent\textbf{VL-Adapter.} After the visual encoder, we build a VL-Adapter using an MLP with one hidden layer. The sign-wise features from visual space $R^{N\times C}$ are mapped into textual space $R^{N\times D}$ by the VL-Adapter as follows:
\begin{equation}
    g_{1:N} = \mbox{VL-Adapter}(f_{1:N}). 
\end{equation}

\noindent\textbf{Light-T.} We pick transformer~\cite{vaswani2017attention} as our lightweight translation model. It contains a text encoder and a text decoder which are composed of several transformer layers. The input features are added with a positional encoding (PE) as $ \hat{g}_n  = g_n + PE(n)$. The text encoder models the global timing information of the input as follows:
\begin{equation}
    h_{1:N} = \mbox{TextEncoder}(\hat{g}_{1:N}). 
\end{equation}
Meanwhile, the corresponding spoken sentence $S = (w_1,w_2,...w_U)$ is tokenized into $S^{'} = (o_1,o_2,...o_L)$ by the same tokenizer from the LLM we will use next. Then it is passed through the word embedding layer (WEL) and a positional encoding (PE) layer as:
\begin{equation}
    z_{i} = \mbox{WEL}(o_i) + \mbox{PE}(i).
\end{equation}
 The text decoder takes word embeddings alone with sign hidden states $h_{1:N}$ as input to generate a predicted sentence one word at a time:
\begin{equation}
    y_{i} = \mbox{TextDecoder}(z_{1:i-1},h_{1:N}).
\end{equation}
A language modeling head (Lm\_Head) is plugged after the text decoder to calculate the conditional probabilities as follows:
\begin{equation}
   p(o_{i}|o_{1:i-1},V) =  (\mbox{softmax}(\mbox{Lm\_Head}(y_{i})))_{o_{i}}.
\end{equation}

\noindent\textbf{Training.} 
We train the model using the video-grounded text generation objective, which aims to generate spoken sentences corresponding to the input videos.
We use the ground truth spoken sentences to calculate the cross-entropy loss and optimize the entire network:
\begin{equation}
    \mathcal {L}_{CE} = - \sum\limits_{i=1}^{L} \mbox {log}\ p(o_{i}|o_{1:i-1},V). \label{loss}
\end{equation}

After the above workflow, we finished initializing the visual encoder on the sign language datasets. The well-initialized visual encoder is now capable of extracting text-oriented features from sign videos. Next, we will take advantage of LLM to generate more approximate and fluent spoken sentences.

\subsection{LLM Fine-tuning}
\label{LLM}
Now we present how the proposed method can exploit the potential of LLM in Gloss-free SLT. In general, we keep the pre-trained visual encoder and plug it into an LLM-Adapter and an LLM. During training, the visual encoder is frozen while the other modules are fine-tuned. 

\noindent \textbf{LLM Selection.} 
The LLM selection follows two standards. Firstly, the selected LLM should be an encoder-decoder architecture because SLT is a translation-type downstream task. Secondly, LLM pre-trained on multilingual corpus is preferred because different datasets have spoken sentences in various languages, such as German in PHOENIX14T~\cite{camgoz2018neural} and Chinese in CSL-Daily~\cite{zhou2021improving}. Under these two criteria, we choose MBart~\cite{liu2020multilingual} as our LLM. MBart is a sequence-to-sequence denoising auto-encoder pre-trained on large-scale monolingual corpora in many languages. It has a standard transformer~\cite{vaswani2017attention} architecture with 12 layers of the encoder and 12 layers of the decoder. MBart primarily intends for translation tasks and has been proven to significantly improve the performance of SLT~\cite{chen2022simple}. Although the number of parameters in MBart is only 680M, to the best of our knowledge, it is the largest language model in the SLT domain.

\noindent \textbf{Fine-tuning.} 

The word embedding layer of the MBart's encoder is replaced by an LLM-Adapter. It is simply implemented as
a fully connected MLP with one hidden layer whose output dimension fits with the LLM. We use the well-initialized visual encoder to extract the sign-wise features of the input sign videos and feed them into the LLM-Adapter to generate sign embeddings. Then the MBart's encoder takes sign embeddings as input to generate hidden states. The corresponding spoken sentences are tokenized by a tokenizer and project one-hot vectors into dense text embeddings via MBart’s pre-trained word embedding layer. MBart’s decoder takes the hidden states alone with the text embeddings to generate predicted sentences one word at a time. During fine-tuning, we freeze the visual encoder and optimize other modules using sequence-to-sequence cross-entropy loss as shown in Equation~\ref{loss}.

\section{Experiments}
\subsection{Datasets and Evaluation Metrics}
\noindent \textbf{Datasets.} The experiments are performed on PHOENIX14T~\citeplanguageresource{camgoz2018neural}, CSL-Daily~\citeplanguageresource{zhou2021improving} and How2Sign~\citeplanguageresource{duarte2021how2sign}. PHOENIX14T is a German Sign Language (DGS) dataset taken from a TV broadcast whose topic focuses
on weather forecasts. It contains 7096, 519, and 642 video-gloss-text pairs in train, dev, and test set, respectively. CSL-Daily is a Chinese Sign Language (CSL) dataset that contains 18401, 1077, and 1176 video-gloss-text pairs in train, dev, and test set, respectively. It is recorded in the laboratory whose topic focuses on daily life. How2Sign is an American Sign Language (ASL) dataset that contains 31164, 1740, and 2356 video-text pairs in train, dev, and test set, respectively. It is recorded in the laboratory and focuses on instructional topics corresponding to various categories. 
The proposed method is compared with state-of-the-art methods on three datasets and conducted ablation analysis on PHOENIX14T. 
We \mbox{report all the results on the test set.}

\noindent \textbf{Protocol.}
Our experiments follow \textit{Gloss-free Sign2Text} protocol proposed by ~\cite{lin2023gloss}. It requests a direct translation from sign language videos to the corresponding spoken sentences without gloss assistance through the entire framework.

\noindent \textbf{Evaluation Metrics.} Following ~\cite{zhou2021improving,chen2022twostream,lin2023gloss,yin2023gloss}, we adopt ROUGE-L~\cite{lin2004rouge} and BLEU~\cite{papineni2002bleu} to evaluate SLT performance.

\begin{table*}[t]
  \centering
  \resizebox{\linewidth}{!}{
  \begin{tabular}{@{}l|c|ccccc@{}}
  \hline
    
    { Method} & Gloss-Free  &   Rouge-L &  BLEU-1 &  BLEU-2 &  BLEU-3 &  BLEU-4 \\
    \hline
    
     SLRT~\cite{camgoz2020sign} &$\times$ & - & 46.61 & 33.73 & 26.19 & 21.32 \\
     STMC-T~\cite{zhou2021spatial} &$\times$  & 46.65& 46.98 & 36.09 & 28.70 & 23.65 \\
     SignBT~\cite{zhou2021improving} &$\times$  & 49.54 & 50.80 & 37.75 & 29.72 & 24.32 \\
     MMTLB~\cite{chen2022simple}  &$\times$ & 52.65 & 53.97 & 41.75 & 33.84 & 28.39 \\
     TS-SLT~\cite{chen2022twostream}& $\times$& \textbf{53.48} &  \textbf{54.90} & \textbf{42.43} & \textbf{34.46} & \textbf{28.95} \\
     \hline
     NSLT~\cite{camgoz2018neural} &\checkmark  & 31.80 & 32.24 & 19.03 & 12.83 & 9.58  \\
     SLRT-GF$^*$~\cite{camgoz2020sign}  &\checkmark   & 31.10 & 30.88 & 18.57 & 13.12 & 10.19 \\
     TK-SLT~\cite{orbay2020neural}&\checkmark   & 36.28 & 37.22 & 23.88 & 17.08 & 13.25\\
     TSPNet~\cite{li2020tspnet} &\checkmark  &  34.96 & 36.10  & 23.12 & 16.88 & 13.41  \\
     CSGCR~\cite{zhao2021conditional} &\checkmark & 38.85 & 36.71 & 25.40 & 18.86 & 15.18  \\
     GASLT~\cite{yin2023gloss} &\checkmark & 39.86 & 39.07  & 26.74 & 21.86 & 15.74  \\
    GFSLT-VLP~\cite{Zhou_2023_ICCV}  &\checkmark  & 42.49 & 43.71 & 33.18 & 26.11 & 21.44 \\
     \hline
      \textbf{FLa-LLM(ours)} & \checkmark  & \textbf{45.27} & \textbf{46.29} & \textbf{35.33} & \textbf{28.03} & \textbf{23.09}\\
      \textcolor{blue}{Improvement} & & \textcolor{blue}{+2.78} & \textcolor{blue}{+2.58} & \textcolor{blue}{+2.15} & \textcolor{blue}{+1.92}& \textcolor{blue}{+1.65}\\
    \hline
  \end{tabular}}
  \caption{Experimental results on PHOENIX14T dataset. * denotes methods reproduced by~\cite{yin2023gloss}. We bold the best results in the gloss-based setting and gloss-free setting. \textcolor{blue}{Improvement} represents comparisons with the previous best gloss-free result.}
  \label{tab:PHOENIX14T}
\end{table*}

\subsection{Implementation Details}
\label{setting}

\begin{table*}[!htp]
  \centering
  \resizebox{\linewidth}{!}{
  \begin{tabular}{@{}l|c|ccccc@{}}
    \hline
    { Method} & Gloss-Free  &   Rouge-L &  BLEU-1 &  BLEU-2 &  BLEU-3 &  BLEU-4 \\
    \hline
    
     SLRT$^\dag$~\cite{camgoz2020sign} &$\times$  & 36.74 & 37.38 & 24.36 & 16.55 & 11.79 \\
     SignBT~\cite{zhou2021improving} & $\times$  & 49.31 & 51.42 & 37.26 & 27.76 & 21.34 \\
    MMTLB~\cite{chen2022simple} &$\times$  & 53.25 & 53.31 & 40.41 & 30.87 & 23.92 \\
     TS-SLT~\cite{chen2022twostream}& $\times$ & \textbf{55.72} &  \textbf{55.44} & \textbf{42.59} & \textbf{32.87} & \textbf{25.79} \\
    \hline
     NSLT$^\dag$~\cite{camgoz2018neural} &\checkmark & 34.54 & 34.16 & 19.57 & 11.84 & 7.56  \\
     TSPNet$^*$~\cite{li2020tspnet} &\checkmark & 18.38 & 17.09  & 8.98 & 5.07 & 2.97  \\
     GASLT~\cite{yin2023gloss} &\checkmark &  20.35 & 19.90  & 9.94& 5.98 & 4.07  \\
     GFSLT-VLP~\cite{Zhou_2023_ICCV} &\checkmark  & 36.44 & \textbf{39.37} & 24.93 & 16.26 & 11.00 \\
     \hline
      \textbf{FLa-LLM(ours)} &\checkmark & \textbf{37.25} & 37.13 & \textbf{25.12} & \textbf{18.38} & \textbf{14.20}\\
      \textcolor{blue}{Improvement} & & \textcolor{blue}{+0.81} & \textcolor{blue}{-2.24} & \textcolor{blue}{+0.19} & \textcolor{blue}{+2.12}& \textcolor{blue}{+3.20}\\
    \hline
  \end{tabular}}
  \caption{Experimental results on CSL-daily dataset. * denotes methods reproduced by~\cite{yin2023gloss}. \dag \ denotes methods reproduced by~\cite{zhou2021improving}. We bold the highest scores in the gloss-based setting and gloss-free setting. \textcolor{blue}{Improvement} represents comparisons with the previous best gloss-free result.}
  \label{tab:CSL-Daily}
\end{table*}

Our model is implemented using the Pytorch framework~\cite{paszke2019pytorch}. The experiments are conducted on NVIDIA GeForce RTX 3090 GPUs. 




\noindent \textbf{Network setting.} We choose ResNet18~\cite{he2016deep} pretrained on ImageNet~\cite{deng2009imagenet} as visual backbone. The local temporal module uses a combination of Conv1D-BN-Relu layers.
The Light-T has 3 transformer layers for both encoder and decoder.
Each layer has an attention head of 8, a hidden size of 512, and a feed-forward dim of 2048. Our LLM is initialized with the official release of MBart-large-cc25\footnote{\url{https://huggingface.co/facebook/MBart-large-cc25}}. The model and corresponding tokenizers are trimmed using the translation corpus of the target SLT train set to save GPU memory.

\noindent \textbf{Training and Inference.} In the visual initialing stage, the model is trained using SGD optimizer~\cite{robbins1951stochastic} with 0.9 momentum and a batch size of 8 across 2 GPUs. The learning rate is set to $1 \times 10^{-2}$ with a cosine annealing schedule~\cite{loshchilov2016sgdr}. In the LLM fine-tuning stage, the model is trained using Adam optimizer~\cite{kingma2014adam} with a batch size of 16 on 1 GPU. The learning rate is set to $1 \times 10^{-5}$ for the LLM and $1 \times 10^{-3}$ for the LLM-Adapter layer with a cosine annealing schedule. We employ cross-entropy loss with a label smoothing of 0.2 in both stages. During inference, we use beam search strategy~\cite{wu2016google} with a beam size of 5.

\subsection{Comparison with State-of-the-art Methods}

\noindent \textbf{Results on PHOENIX14T dataset.} We compare our method with state-of-the-art gloss-based and gloss-free SLT approaches in Table~\ref{tab:PHOENIX14T}. With the gloss-free setting, our method achieves a significant breakthrough in all metrics compared to the previous methods. In particular, we get an outstanding BLEU-4 improvement of 1.65 on the test set compared with the previous state-of-the-art method GFSLT-VLP~\cite{Zhou_2023_ICCV}. Surprisingly, our approach is fairly comparable to some gloss-based approaches, such as SLRT~\cite{camgoz2020sign}, STMC-T~\cite{zhou2021spatial} and SignBT~\cite{zhou2021improving}. 
The performance of our method is still far from MMTLB~\cite{chen2022simple} and TS-SLT~\cite{chen2022twostream}, which also utilizes the LLM capability to enhance SLT. However, they rely heavily on gloss for visual and linguistic pre-training with great limitations.

\begin{table*}[!htp]
  \centering
  \resizebox{\linewidth}{!}{
  \begin{tabular}{@{}l|c|ccccc@{}}
   
    \hline
    { Method} & Gloss-Free  &   Rouge-L &  BLEU-1 &  BLEU-2 &  BLEU-3 &  BLEU-4 \\
    \hline
    
     TF-H2S~\cite{alvarezsign} &\checkmark & - & 17.40 & 7.69 & 3.97 & 2.21  \\
     SLT-IV~\cite{tarres2023sign} &\checkmark & - & \textbf{34.01}  & \textbf{19.30} & 12.18 & 8.03  \\
     GloFE-VN~\cite{lin2023gloss}&\checkmark & 12.61 & 14.94  & 7.27 & 3.93 & 2.24 \\
     \hline
      \textbf{FLa-LLM(ours)} &\checkmark & \textbf{27.81} & 29.81 & 18.99 & \textbf{13.27} & \textbf{9.66}\\
      \textcolor{blue}{Improvement} & & \textcolor{blue}{+15.20} & \textcolor{blue}{-4.20} & \textcolor{blue}{-0.31} & \textcolor{blue}{+1.09}& \textcolor{blue}{+1.63}\\
    \hline
    
  \end{tabular}}
  \caption{Experimental results on How2Sign dataset. We bold the highest scores. \textcolor{blue}{Improvement} represents comparisons with the previous best gloss-free result.}
  \label{tab:How2Sign}
\end{table*}

\noindent \textbf{Results on CSL-Daily dataset.} Table~\ref{tab:CSL-Daily} shows the comparisons between our method and other state-of-the-art methods on the CSL-Daily dataset. When compared with other gloss-free methods, our method achieves a substantial improvement with a margin of 3.20 in BLEU-4 which is 29.09\% higher than the previous state-of-the-art method GFSLT-VLP~\cite{Zhou_2023_ICCV}. However, there is a big gap between our method and the gloss-based methods. This may be due to the size of the sign word's vocabulary. CSL-daily has more than 2K sign words' vocabulary size resulting in more reliance on glosses.

\noindent \textbf{Results on How2Sign dataset.} In Table~\ref{tab:How2Sign}, our method is compared with other state-of-the-art methods on the How2Sign dataset. The performance of our method is substantially better than TF-H2S~\cite{alvarezsign} and GloFE-VN~\cite{lin2023gloss}. However, we only surpass SLT-IV~\cite{tarres2023sign} on BLEU-3 and BLEU-4 while falling behind on BLEU-1 and BLEU-2. Higher BLEU-3 and BLEU-4 indicate our model has better short phrase generating ability which possibly gains from LLM.
\subsection{Ablation Studies}
Our ablation experiments are performed on the PHOENIX14T test set since it is the most widely used benchmark for SLT. Note that we use R to represent ROUGE-L and B1-B4 to represent BLEU1-BLEU4.

\subsubsection{Ablation on Factorized Learning}
\noindent\textbf{Effect of Factorized Learning Strategy.}
We first verify the effectiveness of our proposed factorized learning strategy. The most straightforward approach is to compare our factorized learning with the end-to-end joint training of the visual encoder and LLM. As shown in Table~\ref{tab:Factorized Learning}, our factorized learning strategy substantially outperforms the end-to-end training approach. This may be because the LLM dominates during end-to-end training as mentioned in Section~\ref{sec:dominance}, resulting in weak supervision for the visual encoder. 
\begin{table}[t]
    \centering
    \resizebox{\linewidth}{!}{
    \begin{tabular}{c|ccccc}
    \hline
    Factorized    & R    & B1   & B2   & B3   & B4\\
    \hline
    $\times$     & 32.52 & 31.96 & 21.96 & 16.32 & 12.90 \\
    \checkmark   & \textbf{45.27} & \textbf{46.29} & \textbf{35.33} & \textbf{28.03} & \textbf{23.09} \\
    \hline
    \end{tabular}
    }
    \caption{Effect of the proposed factorized learning strategy. The first row represents end-to-end joint training of the visual encoder and LLM.}
    \label{tab:Factorized Learning}
\end{table}

\begin{table}[t]
    \centering
    \resizebox{\linewidth}{!}{
    \begin{tabular}{cc|ccccc}
    \hline
    VIS & LFS & R & B1& B2& B3& B4\\
    \hline
     $\times$          & \checkmark  & 17.33 & 17.64 & 10.51 & 7.37  & 5.62 \\
    \checkmark &    $\times$         & 38.67 & 39.09 & 28.20 & 21.83 & 17.69 \\
    \checkmark & \checkmark  & \textbf{45.27} & \textbf{46.29} & \textbf{35.33} & \textbf{28.03} & \textbf{23.09} \\
    
    \hline
    \end{tabular}
    }
    \caption{Effect of each stage. VIS represents the visual initialing stage and LFS means the LLM fine-tuning stage. The first row represents freezing the vision backbone and fine-tuning the other modules.}
    \label{tab:Strategy Portfolio}
\end{table}
\noindent\textbf{Effect of Each Stage.}
In Table~\ref{tab:Strategy Portfolio}, we verify the contribution of each stage in the proposed FLa-LLM method. The first row of the Table~\ref{tab:Strategy Portfolio} means we freeze the vision backbone which is only pre-trained on ImageNet and trained the local temporal module with the LLM-Adapter and LLM end-to-end. It leads to a very poor result without initialing the visual encoder which demonstrates the importance of visual initialing on the sign language datasets. The second row shows the performance of the visual encoder and Light-T i.e. the visual initialing stage performance. The visual initialing stage focuses on the visual encoder resulting in fair performance. Based on sufficient initialization of the visual encoder, we successfully take advantage of the LLM and yield better results as shown in the third row. 

\subsubsection{Ablation on Visual Initializing}

\noindent\textbf{Effect of Downsampling Rate.}
We show the effect of different downsampling rates on the training time and model performance in Table~\ref{tab:downsampling}. When the downsampling rate is not lower than 25\%, it has little impact on the model performance while significantly reducing the training time. Therefore, we choose a sampling rate of 25\% to ensure model performance and save training time.
\begin{table}[t]
    \centering
    \resizebox{\linewidth}{!}{
    \begin{tabular}{l|c|ccccc}
    \hline
    Rate   & Time   & R & B1    & B2    & B3    & B4 \\
    \hline
    100\%  & 17.90h  & 44.55 & 45.68 & 35.01 & 27.82 & 22.96  \\
    50\%   & 9.85h  & 44.60 & 46.22 & 35.12 & 27.90 & 23.04 \\
    25\%   & 4.75h   & \textbf{45.27} & \textbf{46.29} & \textbf{35.33} & \textbf{28.03} & \textbf{23.09} \\
    12.5\% & 3.55h   & 40.77 & 42.42 & 31.62 & 24.68 & 20.02 \\
    \hline
    \end{tabular}
    }
    \caption{Effect of downsampling rate. The second column represents the time required to complete the visual initialing stage.}
    \label{tab:downsampling}
\end{table}

\noindent\textbf{Effect of Light-T.} 
We investigate whether the scale of the Light-T in the visual initialing stage affects the final LLM fine-tuning results. As shown in Table~\ref{tab:Lightweight trans}, the transformer scale has little impact on final performance. This indicates that the main focus during the visual initialing stage is on the visual encoder. The visual encoder can get a good sign language representation ability after initialing regardless of the transformer scale connected.
\begin{table}[t]
    \centering
    \resizebox{\linewidth}{!}{
    \begin{tabular}{l|c|c|c}
    \hline
    Size & Settings & Params &B4\\
    \hline
    Tiny & (1,4,256,1024) & 3.61M & 22.52\\
    Small & (2,4,512,2048) & 18.25M & 22.36\\
    Base & (3,8,512,2048) & 25.61M & \textbf{23.09}\\
    Large & (4,8,1024,4096) & 124.66M &  22.49 \\
    \hline
    \end{tabular}
    }
    \caption{Effect of translation network scale. The (1,4,256,1024) in the second column represents that the transformer has 1 hidden layer, 4 attention heads, a hidden size of 256, and a feed-forward dim of 2048. Params represents the number of model parameters.}
    \label{tab:Lightweight trans}
\end{table}

\begin{figure}[t]
\centering
\includegraphics[width=1.0\linewidth]{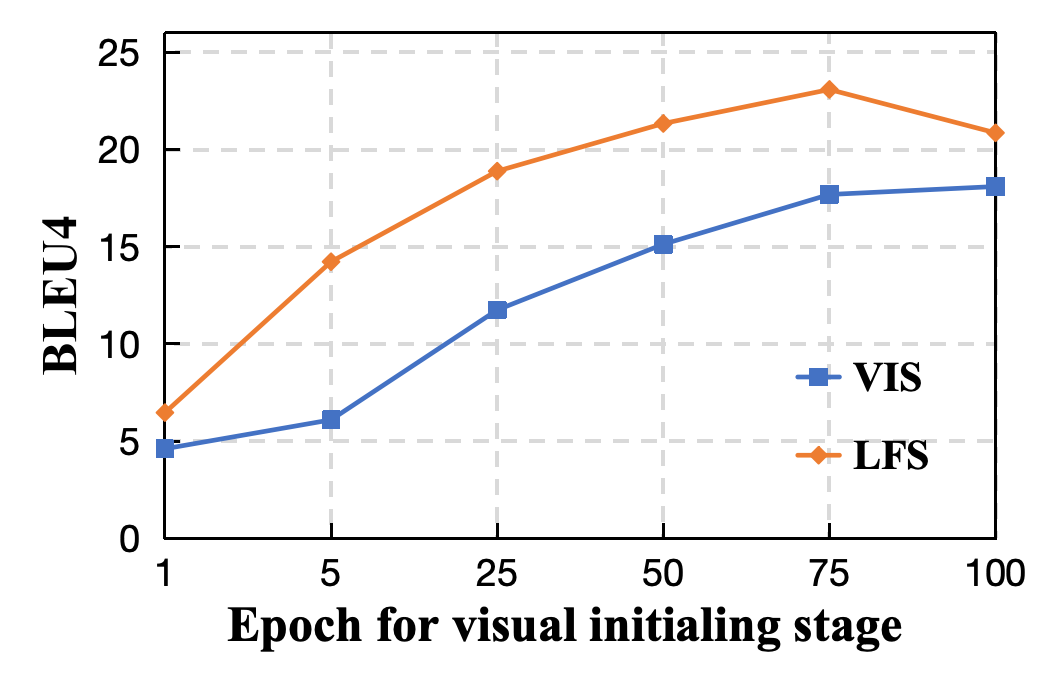}
\caption{Effect of initialing time. VIS represents the visual initialing stage and LFS represents the LLM fine-tuning stage.}
\label{fig:epoch}
\end{figure}

\noindent\textbf{Effect of Initialing Time.}
 We train various visual encoders with different epochs in the initialing stage and use them to do LLM fine-tuning in the same setting. Figure~\ref{fig:epoch} shows the results of the visual initialing stage and the LLM fine-tuning stage with different initialing epochs. LLM fine-tuning can significantly improve the model translation capability in all cases. When the initialing epoch is 100, the performance of LLM fine-tuning decreases compared to epoch 75, probably due to model overfitting. Therefore, we choose the model with epoch 75 for fine-tuning in other experiments.

\subsubsection{Ablation on LLM Fine-tuning}
\noindent\textbf{Effect of Input Features.}
In Table~\ref{tab:features}, we investigate the effect of different input features of the LLM-Adapter layer. We select three features as shown in Figure~\ref{fig:Overview}, namely frame-wise features, sign-wise features, and sign hidden states. The best result is obtained by using sign-wise features as input for LLM fine-tuning. This may be due to the fact that sign-wise features contain both spatial representations and local timing information of sign language videos. 

\begin{table}[t]
    \centering
    \resizebox{\linewidth}{!}{
    \begin{tabular}{l|ccccc}
    \hline
    Features      & R     & B1    & B2    & B3    & B4 \\
    \hline
    Frame-wise    & 41.59 & 42.54 & 31.72 & 24.82 & 20.30  \\
    Sign-wise     & \textbf{45.27} & \textbf{46.29} & \textbf{35.33} & \textbf{28.03} & \textbf{23.09} \\
    Hidden states  & 45.16 & 45.41 & 34.74 & 27.47 & 22.60 \\
    \hline
    \end{tabular}
    }
    \caption{Effect of different input features of LLM. The different features are shown in Figure~\ref{fig:Overview}.}
    \label{tab:features}
\end{table}

\begin{table}[!t]
    \centering
    \resizebox{\linewidth}{!}{
    \begin{tabular}{cc|ccccc}
    \hline
    VB         &  TM        & R     &  B1   & B2    & B3    & B4\\
    \hline
    $\times$       &    $\times$     & 40.72 & 40.74 & 29.79 & 22.65  & 17.86 \\
    \checkmark & $\times$ & 39.86 & 41.64 & 31.48 & 24.86 & 20.40 \\
    \checkmark & \checkmark & \textbf{45.27} & \textbf{46.29} & \textbf{35.33} & \textbf{28.03} & \textbf{23.09}\\
    \hline
    \end{tabular}
    }
    \caption{Effect of freezing different parts of the visual encoder. VB means visual backbone. TM represents the local temporal module. $\checkmark$ means freezing the module while $\times$ means no freezing.}
    \label{tab:freeze}
\end{table}

\begin{table}[htp!]
    \centering
    \resizebox{\linewidth}{!}{
    \begin{tabular}{l|ccccc}
    \hline
    LLM              & R     & B1    & B2    & B3    & B4 \\
    \hline
    MBart w/o pre    & 40.18 & 37.18 & 26.99 & 20.54 & 16.19  \\
    MT5-Base w/o pre    & 22.71 & 18.02 & 12.21 & 9.17  & 7.39 \\
    MBart w/ pre     & \textbf{45.27} & \textbf{46.29} & \textbf{35.33} & \textbf{28.03} & \textbf{23.09}  \\
    MT5-Base w/ pre     & 41.06 & 41.96 & 31.20 & 24.24 & 19.71  \\
    \hline
    \end{tabular}
    }
    \caption{Effect of different LLMs. W/o, w/, and pre means without, with, and pretraining, respectively.}
    \label{tab:llms}
\end{table}

\noindent\textbf{Effect of Frozen Blocks.}
The visual encoder with valid initialization is frozen during the LLM fine-tuning stage. We examined the effect of freezing different parts of the visual encoder in Table~\ref{tab:freeze}. It can be seen that fine-tuning the visual encoder during the LLM fine-tuning stage hurt the performance substantially. This may be due to that LLM dominates the training process and disrupts the initialization of the visual encoder. It also indicates that the visual encoder already has a sufficient visual representation of sign language after the visual initializing stage.

\noindent\textbf{Effect of Different LLMs.}
In order to verify the robustness of our training strategy and to select the most suitable LLM for sign language, we perform the fine-tuning stage using different LLMs. We select two popular multilingual unsupervised pre-training models which are MBart~\cite{liu2020multilingual} and MT5-Base~\cite{xue-etal-2021-mt5}. Table ~\ref{tab:llms} shows the fine-tuning results of these two LLMs with random initialization and after pre-training. It can be seen that unsupervised pre-training on a large-scale corpus can significantly improve the performance of LLM fine-tuning on sign language datasets. In addition, MBart performs better compared to MT5-Base, so we chose MBart as our default LLM.

\section{Conclusion}
In this paper, we propose a factorized learning strategy to transfer LLM for gloss-free SLT. In the visual initialing stage, we use a lightweight translation model to pre-train the visual encoder without gloss supervision. In the LLM fine-tuning stage, we freeze the well-initialized visual encoder and fine-tune a powerful LLM to adapt to the downstream SLT task. By splitting the training into two stages, we avoid performance degradation and utilize LLM in a resource-friendly situation. Our method significantly boosts the performance of gloss-free SLT on several datasets.

\section{Limitations}
Our proposed method has two main drawbacks. First, though the factorized learning strategy can avoid performance degradation, it is more cumbersome compared to end-to-end training. A more ideal approach would be adding additional constraints on the visual encoder to the end-to-end framework. Second, in the fine-tuning stage, we fine-tune all parameters of the large language model which limits the scale of our LLM. In future work, we will investigate parameter-efficient fine-tuning methods such as Lora~\cite{hu2022lora} and Prefix-Turning~\cite{li2021prefix}.

\section{Acknowledgements}
This work was supported by the National Key Research and Development Program of China under Grant 2021YFE0205700, Beijing Natural Science Foundation JQ23016, the External cooperation key project of Chinese Academy Sciences 173211KYSB20200002, the Science and Technology Development Fund of Macau Project 0123/2022/A3, and 0070/2020/AMJ, Open Research Projects of Zhejiang Lab No. 2021KH0AB07, and CCF-Zhipu AI Large Model Project 202219.

\section{Bibliographical References}\label{sec:reference}

\bibliographystyle{lrec-coling2024-natbib}
\bibliography{lrec-coling2024-example}

\begin{thebibliography}{3}
\expandafter\ifx\csname natexlab\endcsname\relax\def\natexlab#1{#1}\fi

\bibitem[{Camgoz et~al.(2018)Camgoz, Hadfield, Koller, Ney, and Bowden}]{camgoz2018neural}
Camgoz, Necati Cihan and Hadfield, Simon and Koller, Oscar and Ney, Hermann and Bowden, Richard. 2018.
\newblock \emph{RWTH-PHOENIX-2014-T}.
\newblock PID \href{https://www-i6.informatik.rwth-aachen.de/~koller/RWTH-PHOENIX-2014-T/}{https://www-i6.informatik.rwth-aachen.de/~koller/RWTH-PHOENIX-2014-T/}.

\bibitem[{Duarte et~al.(2021)Duarte, Palaskar, Ventura, Ghadiyaram, DeHaan, Metze, Torres, and Giro-i Nieto}]{duarte2021how2sign}
Duarte, Amanda and Palaskar, Shruti and Ventura, Lucas and Ghadiyaram, Deepti and DeHaan, Kenneth and Metze, Florian and Torres, Jordi and Giro-i-Nieto, Xavier. 2021.
\newblock \emph{How2sign}.
\newblock PID \href{https://how2sign.github.io/}{https://how2sign.github.io/}.

\bibitem[{Zhou et~al.(2021)Zhou, Zhou, Qi, Pu, and Li}]{zhou2021improving}
Zhou, Hao and Zhou, Wengang and Qi, Weizhen and Pu, Junfu and Li, Houqiang. 2021.
\newblock \emph{CSL-Daily}.
\newblock PID \href{https://ustc-slr.github.io/datasets/2021\_csl\_daily/}{https://ustc-slr.github.io/datasets/2021\_csl\_daily/}.

\end{thebibliography}


\begin{thebibliography}{32}
\expandafter\ifx\csname natexlab\endcsname\relax\def\natexlab#1{#1}\fi

\bibitem[{Alvarez et~al.()Alvarez, Nieto, and Benet}]{alvarezsign}
Patricia~Cabot Alvarez, Xavier~Gir{\'o} Nieto, and Laia~Tarr{\'e}s Benet.
\newblock Sign language translation based on transformers for the how2sign dataset.

\bibitem[{Bengio et~al.(2000)Bengio, Ducharme, and Vincent}]{bengio2000neural}
Yoshua Bengio, R{\'e}jean Ducharme, and Pascal Vincent. 2000.
\newblock A neural probabilistic language model.
\newblock \emph{Advances in neural information processing systems}, 13.

\bibitem[{Camgoz et~al.(2018)Camgoz, Hadfield, Koller, Ney, and Bowden}]{camgoz2018neural}
Necati~Cihan Camgoz, Simon Hadfield, Oscar Koller, Hermann Ney, and Richard Bowden. 2018.
\newblock Neural sign language translation.
\newblock In \emph{Proceedings of the IEEE conference on computer vision and pattern recognition}, pages 7784--7793.

\bibitem[{Camgoz et~al.(2020)Camgoz, Koller, Hadfield, and Bowden}]{camgoz2020sign}
Necati~Cihan Camgoz, Oscar Koller, Simon Hadfield, and Richard Bowden. 2020.
\newblock Sign language transformers: Joint end-to-end sign language recognition and translation.
\newblock In \emph{Proceedings of the IEEE/CVF conference on computer vision and pattern recognition}, pages 10023--10033.

\bibitem[{Chen et~al.(2022{\natexlab{a}})Chen, Wei, Sun, Wu, and Lin}]{chen2022simple}
Yutong Chen, Fangyun Wei, Xiao Sun, Zhirong Wu, and Stephen Lin. 2022{\natexlab{a}}.
\newblock A simple multi-modality transfer learning baseline for sign language translation.
\newblock In \emph{Proceedings of the IEEE/CVF Conference on Computer Vision and Pattern Recognition}, pages 5120--5130.

\bibitem[{Chen et~al.(2022{\natexlab{b}})Chen, Zuo, Wei, Wu, LIU, and Mak}]{chen2022twostream}
Yutong Chen, Ronglai Zuo, Fangyun Wei, Yu~Wu, Shujie LIU, and Brian Mak. 2022{\natexlab{b}}.
\newblock \href {https://openreview.net/forum?id=hSxK-4KGLbI} {Two-stream network for sign language recognition and translation}.
\newblock In \emph{Advances in Neural Information Processing Systems}.

\bibitem[{Deng et~al.(2009)Deng, Dong, Socher, Li, Li, and Fei-Fei}]{deng2009imagenet}
Jia Deng, Wei Dong, Richard Socher, Li-Jia Li, Kai Li, and Li~Fei-Fei. 2009.
\newblock Imagenet: A large-scale hierarchical image database.
\newblock In \emph{2009 IEEE conference on computer vision and pattern recognition}, pages 248--255. Ieee.

\bibitem[{Duarte et~al.(2021)Duarte, Palaskar, Ventura, Ghadiyaram, DeHaan, Metze, Torres, and Giro-i Nieto}]{duarte2021how2sign}
Amanda Duarte, Shruti Palaskar, Lucas Ventura, Deepti Ghadiyaram, Kenneth DeHaan, Florian Metze, Jordi Torres, and Xavier Giro-i Nieto. 2021.
\newblock How2sign: a large-scale multimodal dataset for continuous american sign language.
\newblock In \emph{Proceedings of the IEEE/CVF conference on computer vision and pattern recognition}, pages 2735--2744.

\bibitem[{He et~al.(2016)He, Zhang, Ren, and Sun}]{he2016deep}
Kaiming He, Xiangyu Zhang, Shaoqing Ren, and Jian Sun. 2016.
\newblock Deep residual learning for image recognition.
\newblock In \emph{Proceedings of the IEEE conference on computer vision and pattern recognition}, pages 770--778.

\bibitem[{Hu et~al.(2022)Hu, Shen, Wallis, Allen-Zhu, Li, Wang, Wang, and Chen}]{hu2022lora}
Edward~J Hu, Yelong Shen, Phillip Wallis, Zeyuan Allen-Zhu, Yuanzhi Li, Shean Wang, Lu~Wang, and Weizhu Chen. 2022.
\newblock \href {https://openreview.net/forum?id=nZeVKeeFYf9} {Lo{RA}: Low-rank adaptation of large language models}.
\newblock In \emph{International Conference on Learning Representations}.

\bibitem[{Kingma and Ba(2014)}]{kingma2014adam}
Diederik~P Kingma and Jimmy Ba. 2014.
\newblock Adam: A method for stochastic optimization.
\newblock \emph{arXiv preprint arXiv:1412.6980}.

\bibitem[{Li et~al.(2020)Li, Xu, Yu, Zhang, Swift, Suominen, and Li}]{li2020tspnet}
Dongxu Li, Chenchen Xu, Xin Yu, Kaihao Zhang, Benjamin Swift, Hanna Suominen, and Hongdong Li. 2020.
\newblock Tspnet: Hierarchical feature learning via temporal semantic pyramid for sign language translation.
\newblock \emph{Advances in Neural Information Processing Systems}, 33:12034--12045.

\bibitem[{Li and Liang(2021)}]{li2021prefix}
Xiang~Lisa Li and Percy Liang. 2021.
\newblock Prefix-tuning: Optimizing continuous prompts for generation.
\newblock In \emph{Proceedings of the 59th Annual Meeting of the Association for Computational Linguistics and the 11th International Joint Conference on Natural Language Processing (Volume 1: Long Papers)}, pages 4582--4597.

\bibitem[{Lin(2004)}]{lin2004rouge}
Chin-Yew Lin. 2004.
\newblock Rouge: A package for automatic evaluation of summaries.
\newblock In \emph{Text summarization branches out}, pages 74--81.

\bibitem[{Lin et~al.(2023)Lin, Wang, Zhu, Sun, Zhang, and Yang}]{lin2023gloss}
Kezhou Lin, Xiaohan Wang, Linchao Zhu, Ke~Sun, Bang Zhang, and Yi~Yang. 2023.
\newblock Gloss-free end-to-end sign language translation.
\newblock \emph{arXiv preprint arXiv:2305.12876}.

\bibitem[{Liu et~al.(2020)Liu, Gu, Goyal, Li, Edunov, Ghazvininejad, Lewis, and Zettlemoyer}]{liu2020multilingual}
Yinhan Liu, Jiatao Gu, Naman Goyal, Xian Li, Sergey Edunov, Marjan Ghazvininejad, Mike Lewis, and Luke Zettlemoyer. 2020.
\newblock Multilingual denoising pre-training for neural machine translation.
\newblock \emph{Transactions of the Association for Computational Linguistics}, 8:726--742.

\bibitem[{Loshchilov and Hutter(2016)}]{loshchilov2016sgdr}
Ilya Loshchilov and Frank Hutter. 2016.
\newblock Sgdr: Stochastic gradient descent with warm restarts.
\newblock \emph{arXiv preprint arXiv:1608.03983}.

\bibitem[{Orbay and Akarun(2020)}]{orbay2020neural}
Alptekin Orbay and Lale Akarun. 2020.
\newblock Neural sign language translation by learning tokenization.
\newblock In \emph{2020 15th IEEE International Conference on Automatic Face and Gesture Recognition (FG 2020)}, pages 222--228. IEEE.

\bibitem[{Papineni et~al.(2002)Papineni, Roukos, Ward, and Zhu}]{papineni2002bleu}
Kishore Papineni, Salim Roukos, Todd Ward, and Wei-Jing Zhu. 2002.
\newblock Bleu: a method for automatic evaluation of machine translation.
\newblock In \emph{Proceedings of the 40th annual meeting of the Association for Computational Linguistics}, pages 311--318.

\bibitem[{Paszke et~al.(2019)Paszke, Gross, Massa, Lerer, Bradbury, Chanan, Killeen, Lin, Gimelshein, Antiga et~al.}]{paszke2019pytorch}
Adam Paszke, Sam Gross, Francisco Massa, Adam Lerer, James Bradbury, Gregory Chanan, Trevor Killeen, Zeming Lin, Natalia Gimelshein, Luca Antiga, et~al. 2019.
\newblock Pytorch: An imperative style, high-performance deep learning library.
\newblock \emph{Advances in neural information processing systems}, 32.

\bibitem[{Robbins and Monro(1951)}]{robbins1951stochastic}
Herbert Robbins and Sutton Monro. 1951.
\newblock A stochastic approximation method.
\newblock \emph{The annals of mathematical statistics}, pages 400--407.

\bibitem[{Shi et~al.(2022)Shi, Brentari, Shakhnarovich, and Livescu}]{shi2022open}
Bowen Shi, Diane Brentari, Greg Shakhnarovich, and Karen Livescu. 2022.
\newblock Open-domain sign language translation learned from online video.
\newblock In \emph{EMNLP}.

\bibitem[{Tarr{\'e}s et~al.(2023)Tarr{\'e}s, G{\'a}llego, Duarte, Torres, and Gir{\'o}-i Nieto}]{tarres2023sign}
Laia Tarr{\'e}s, Gerard~I G{\'a}llego, Amanda Duarte, Jordi Torres, and Xavier Gir{\'o}-i Nieto. 2023.
\newblock Sign language translation from instructional videos.
\newblock In \emph{Proceedings of the IEEE/CVF Conference on Computer Vision and Pattern Recognition}, pages 5624--5634.

\bibitem[{Vaswani et~al.(2017)Vaswani, Shazeer, Parmar, Uszkoreit, Jones, Gomez, Kaiser, and Polosukhin}]{vaswani2017attention}
Ashish Vaswani, Noam Shazeer, Niki Parmar, Jakob Uszkoreit, Llion Jones, Aidan~N Gomez, {\L}ukasz Kaiser, and Illia Polosukhin. 2017.
\newblock Attention is all you need.
\newblock \emph{Advances in neural information processing systems}, 30.

\bibitem[{Wu et~al.(2016)Wu, Schuster, Chen, Le, Norouzi, Macherey, Krikun, Cao, Gao, Macherey et~al.}]{wu2016google}
Yonghui Wu, Mike Schuster, Zhifeng Chen, Quoc~V Le, Mohammad Norouzi, Wolfgang Macherey, Maxim Krikun, Yuan Cao, Qin Gao, Klaus Macherey, et~al. 2016.
\newblock Google's neural machine translation system: Bridging the gap between human and machine translation.
\newblock \emph{arXiv preprint arXiv:1609.08144}.

\bibitem[{Xue et~al.(2021)Xue, Constant, Roberts, Kale, Al-Rfou, Siddhant, Barua, and Raffel}]{xue-etal-2021-mt5}
Linting Xue, Noah Constant, Adam Roberts, Mihir Kale, Rami Al-Rfou, Aditya Siddhant, Aditya Barua, and Colin Raffel. 2021.
\newblock \href {https://doi.org/10.18653/v1/2021.naacl-main.41} {m{T}5: A massively multilingual pre-trained text-to-text transformer}.
\newblock In \emph{Proceedings of the 2021 Conference of the North American Chapter of the Association for Computational Linguistics: Human Language Technologies}, pages 483--498, Online. Association for Computational Linguistics.

\bibitem[{Yin et~al.(2023)Yin, Zhong, Tang, Jin, Jin, and Zhao}]{yin2023gloss}
Aoxiong Yin, Tianyun Zhong, Li~Tang, Weike Jin, Tao Jin, and Zhou Zhao. 2023.
\newblock Gloss attention for gloss-free sign language translation.
\newblock In \emph{Proceedings of the IEEE/CVF Conference on Computer Vision and Pattern Recognition}, pages 2551--2562.

\bibitem[{Yin et~al.(2021)Yin, Moryossef, Hochgesang, Goldberg, and Alikhani}]{yin2021including}
Kayo Yin, Amit Moryossef, Julie Hochgesang, Yoav Goldberg, and Malihe Alikhani. 2021.
\newblock Including signed languages in natural language processing.
\newblock In \emph{Proceedings of the 59th Annual Meeting of the Association for Computational Linguistics and the 11th International Joint Conference on Natural Language Processing (Volume 1: Long Papers)}, pages 7347--7360.

\bibitem[{Zhao et~al.(2021)Zhao, Qi, Zhou, Duan, Zhou, and Li}]{zhao2021conditional}
Jian Zhao, Weizhen Qi, Wengang Zhou, Nan Duan, Ming Zhou, and Houqiang Li. 2021.
\newblock Conditional sentence generation and cross-modal reranking for sign language translation.
\newblock \emph{IEEE Transactions on Multimedia}, 24:2662--2672.

\bibitem[{Zhou et~al.(2023)Zhou, Chen, Clap\'es, Wan, Liang, Escalera, Lei, and Zhang}]{Zhou_2023_ICCV}
Benjia Zhou, Zhigang Chen, Albert Clap\'es, Jun Wan, Yanyan Liang, Sergio Escalera, Zhen Lei, and Du~Zhang. 2023.
\newblock Gloss-free sign language translation: Improving from visual-language pretraining.
\newblock In \emph{Proceedings of the IEEE/CVF International Conference on Computer Vision (ICCV)}, pages 20871--20881.

\bibitem[{Zhou et~al.(2021{\natexlab{a}})Zhou, Zhou, Qi, Pu, and Li}]{zhou2021improving}
Hao Zhou, Wengang Zhou, Weizhen Qi, Junfu Pu, and Houqiang Li. 2021{\natexlab{a}}.
\newblock Improving sign language translation with monolingual data by sign back-translation.
\newblock In \emph{Proceedings of the IEEE/CVF Conference on Computer Vision and Pattern Recognition}, pages 1316--1325.

\bibitem[{Zhou et~al.(2021{\natexlab{b}})Zhou, Zhou, Zhou, and Li}]{zhou2021spatial}
Hao Zhou, Wengang Zhou, Yun Zhou, and Houqiang Li. 2021{\natexlab{b}}.
\newblock Spatial-temporal multi-cue network for sign language recognition and translation.
\newblock \emph{IEEE Transactions on Multimedia}, 24:768--779.

\end{thebibliography}

\section{Language Resource References}
\label{lr:ref}
\bibliographystylelanguageresource{lrec-coling2024-natbib}
\bibliographylanguageresource{languageresource}

\end{document}